%% file: main.tex
\pdfoutput=1

\documentclass[11pt]{article}

\usepackage[]{emnlp2021}

\usepackage{times}
\usepackage{latexsym}

\usepackage[T1]{fontenc}

\usepackage[utf8]{inputenc}

\usepackage{microtype}

\usepackage{listings}
\usepackage{latexsym}

\usepackage{graphicx}
\usepackage{caption}
\usepackage{subcaption}

\title{DILBERT: Customized Pre-Training for Domain Adaptation with Category Shift, with an Application to Aspect Extraction}

\author{Entony Lekhtman $^1$ Yftah Ziser$^2$  Roi Reichart$^3$ 
     \\
       $^1$Faculty of Computer Science, Technion, IIT  
       \\
       $^2$School of Informatics, University of Edinburgh
 \\
       $^3$Faculty of Industrial Engineering and Management, Technion, IIT  
       \\
       \tt\small tony1@campus.technion.ac.il \tt\small 	yftah.ziser@ed.ac.uk \tt\small roiri@technion.ac.il 
       \\
       }

\date{}

\begin{document}
\maketitle
\begin{abstract}
The rise of pre-trained language models has yielded substantial progress in the vast majority of Natural Language Processing (NLP) tasks. However, a generic approach towards the pre-training procedure can naturally be sub-optimal in some cases. Particularly, fine-tuning a pre-trained language model on a source domain  and then applying it to a different target domain, results in a sharp performance decline of the eventual classifier for many source-target domain pairs. Moreover, in some NLP tasks, the output categories substantially differ between domains, making adaptation even more challenging. This, for example, happens in the task of aspect extraction, where the aspects of interest of reviews of, e.g., restaurants or electronic devices may be very different.  
This paper presents a new fine-tuning scheme for BERT, which aims to address the above challenges. We name this scheme  \textit{DILBERT: Domain Invariant Learning with BERT}, and customize it  for aspect extraction in the unsupervised domain adaptation setting. DILBERT harnesses the categorical information of both the source and the target domains to guide the pre-training process towards a more domain and category invariant representation, thus closing the gap between the domains. 
We show that DILBERT yields substantial improvements over state-of-the-art baselines while using a fraction of the unlabeled data, particularly in more challenging domain adaptation setups.\footnote{Our code is publicly available at: https://github.com/tonylekhtman/DILBERT}

\end{abstract}
\input{intro}
\input{related_work}

\input{method}

\input{exp_setup}

\input{results}

\input{conclusions}
\section*{Acknowledgments}
We would like to thank the members
of the IE@Technion NLP group for their valuable
feedback and advice. This research was partially
funded by an ISF personal grant No. 1625/18.
\bibliographystyle{acl_natbib}
\bibliography{ref}
\input{supp-materail}

\end{document}

%% file: intro.tex
\section{Introduction} \label{sec:intro}

\begin{figure}
   \centering
   \includegraphics[width=3in]{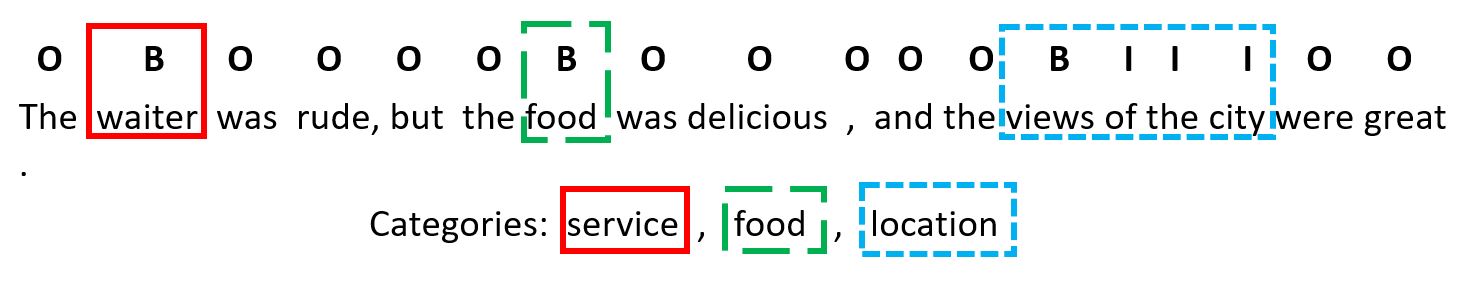}
   \caption{A review from the Restaurants domain. We use BIO tags to mark the spans of the  aspects. Each aspect belongs to one category: For example, the \textit{waiter}  aspect belongs to the \textit{service} category. In this review, the word \textit{food} serves as both an aspect and a category.}
   \label{fig:res_review}
\end{figure}

Aspect-based sentiment analysis (ABSA) \cite{thet2010aspect}, extracting aspect-sentiment pairs  for products or services from reviews, is a widely researched task in both academia and industry. ABSA allows a fine-grained and realistic evaluation of reviews, as real-world reviews typically do not convey a homogeneous sentiment but rather communicate different sentiments for different aspects of the reviewed item or service. For example, while the overall sentiment of the review in Figure~\ref{fig:res_review} is unclear, the sentiment towards the service, food, and location of the restaurant is very decisive. Moreover, even when the overall sentiment of the review is clear, ABSA provides more nuanced and complete information about its content.

In this paper, we focus on the sub-task of aspect extraction (AE, a.k.a opinion targets extraction): Extracting from opinionated texts the aspects on which the reader conveys sentiment. For example, in Figure~\ref{fig:res_review}, the waiter, food, and the views of the city are aspects derived from broader categories: service, food, and location. This task is characterized by a multiplicity of domains, as reviews and other opinionated texts can be written about a variety of products, services as well as many other issues. Moreover, the aspect categories of interest often differ between these domains.

As for most NLP tasks and applications, AE research has recently made substantial progress. While Transformer \cite{DBLP:journals/corr/VaswaniSPUJGKP17} based pre-trained models \cite{devlin-etal-2019-bert, liu2019roberta} have pushed results substantially forward, they rely on in-domain labeled data to achieve their strong results. Annotating such data for multiple domains is costly and laborious, which is one of the major bottlenecks for developing and deploying NLP systems. As noted above, AE forms a particularly challenging variant of the domain adaptation problem, as the aspect categories of interest tend to change across domains.

A well-established approach for addressing the above bottleneck is Domain Adaptation (DA) \cite{blitzer2006domain, ben2007analysis}. DA, training models on source domain labeled data so that they can effectively generalize to different target domains, is a long-standing research challenge. While the target domain labeled data availability in DA setups ranges from little (supervised DA \cite{daume:07}) to none (unsupervised DA \cite{ ramponi2020neural}), unlabeled data is typically available in both source and target domains. This paper focuses on unsupervised DA as we believe it is a more realistic and practical scenario. 

Due to the great success of deep learning models, 
DA through representation learning \cite{blitzer-etal-2007-biographies, ziser-reichart-2017-neural}, i.e., learning a shared representation for both the source and target domains, has recently become prominent \cite{ziser-reichart-2018-deep, ben-david-etal-2020-perl}. Of particular importance to this line of work are approaches that utilize pivot features \cite{blitzer2006domain} that: (a) frequently appear in both domains; and (b) have high mutual  information (MI) with the task label. 
%
%
While pivot-based methods achieve state-of-the-art results in many text classification tasks \cite{ziser-reichart-2018-deep, miller2019simplified, ben-david-etal-2020-perl}, it is not trivial to successfully apply them on tasks such as AE. This stems from two reasons: First, AE is a sequence tagging task with multiple labels for each input example (i.e., word-level labels for input sentences). For a feature to meet the second condition for being a pivot (high MI with the task label), further refinement of the pivot definition is required. Second, different domains often differ in their aspect categories and hence if a feature is highly correlated with a source domain label (aspect category), this is not indicative of its being correlated with the (different) aspect categories of the target domain.

To overcome these limitations, we present \textit{DILBERT: Domain Invariant Learning with BERT}, a customized fine-tuning procedure for AE in an unsupervised DA setup. More specifically, DILBERT employs a variant of the BERT masked language modeling (MLM) task such that hidden tokens are chosen by their semantic similarity to the categories rather than randomly. Further, it employs a new pre-training task: The prediction of which categories appear in the input text. Notice that unlabeled text does not contain supervision for our pre-training tasks and we hence have to use distant supervision as an approximation.

In our unsupervised DA experiments we  consider laptop and restaurant reviews, where for the restaurant domain we consider two variants, that differ in their difficulty. Our best performing model outperforms the strongest baseline by over 5\% on average and over 13\% on the most challenging setup, while using only a small fraction of the unlabeled data. Moreover, we show that our pre-training procedure is very effective in resource-poor setups, where unlabeled data is scarce.

%% file: related_work.tex
\section{Background and Previous Work}
\label{sec:previous}

While both aspect extraction and domain adaptation are active fields of research, research on their intersection is not as frequent. We hence first describe in-domain AE, and then continue with a survey of DA, focusing on pivot-based unsupervised DA. Finally, we  describe works at the intersection of both problems.

\subsection{Aspect Extraction}

Early in-domain aspect extraction works heavily rely  on feature engineering, often feeding graphical models with linguistic-driven features such as part-of-speech (POS) tags \cite{jin2009novel}, WordNet attributes and word frequencies \cite{li-etal-2010-structure}.
The SemEval ABSA task releases  \cite{pontiki-etal-2014-semeval, pontiki-etal-2015-semeval, pontiki-etal-2016-semeval} and the rise of deep learning have pushed AE research substantially forward. 

\newcite{liu-etal-2015-fine} applied Recurrent Neural Networks (RNN) combined with simple linguistic features to outperform feature-rich Conditional Random Field (CRF) models. \newcite{wang-etal-2016-recursive} showed that stacking a CRF on top of an RNN further improves the results. Recently, \newcite{xu2019bert} tuned BERT for AE and obtained additional improvements, demonstrating the effectiveness of massive pre-training with unlabeled data for this task. \citet{tian-etal-2020-skep} proposed new pre-training tasks based on automatically extracted sentiment words and aspect terms.  Finally, \newcite{jiang-etal-2019-challenge} released the Multi-Aspect Multi-Sentiment (MAMS) dataset, where each sentence contains at least two different aspects with different sentiment polarities, making it more challenging compared to the SemEval datasets. 

\subsection{Domain Adaptation}

DA is a fundamental challenge in machine learning in general and NLP in particular. In this work, we focus on unsupervised DA, in which labeled data is available from the source domain and unlabeled data is available from both the source and the target domains. DA approaches include
 instance re-weighting \cite{mansour2008domain, gong-etal-2020-unified}, sub-sampling from both domains \cite{chen2011automatic} and learning a shared representation for the source and target domains \cite{blitzer-etal-2007-biographies, ganin2016domain, ziser-reichart-2018-pivot}. This section focuses on the latter approach which has become prominent due to the success of deep learning. Indeed, our proposed method, as well as the previous methods and the baselines we compare to, follow this approach.  
 
 \paragraph{Unsupervised DA via Shared Representation} 
 The shared representation approach to unsupervised DA typically consists of two major steps: (a) A representation model is trained using the unlabeled data from the source and target domains; and (b) A task-specific classifier is stacked on top of the representation model of step (a) and fine-tuned using the source domain labeled data. The fine-tuned model is then applied to the target domain test data, hoping that the domain-invariant feature space would mitigate the domain gap. Many works have followed this avenue (e.g., \cite{chen2012marginalized,louizos2016variational,ganin2016domain}) and a comprehensive survey is beyond the scope of this paper. Below we discuss the line of work which is most relevant to our model as well as to most previous unsupervised DA for AE work.

\paragraph{Shared Representation Using Pivot Features} 

Pivots features, proposed by \citet{blitzer2006domain,blitzer-etal-2007-biographies} through their structural correspondence learning (SCL)
framework, are features that meet both the domain frequency and the source-domain task label correlation conditions, defined in \S \ref{sec:intro}. The authors use the distinction between pivot and non-pivot features (features that do not meet at least one of the criteria, as long as they are frequent in one of the domains) in order to learn a shared cross-domain representation model. 
The main idea is to utilize the pivot features to extract cross-domain and task-relevant information from non-pivot features, which is done through non-pivot to pivot feature mapping. This way the induced representation consists of the cross-domain and task-relevant information of both feature types.

\citet{blitzer2006domain,blitzer-etal-2007-biographies} learned linear non-pivot to pivot mappings that do not exploit the input structure (e.g., the structure of the review document in a sentiment classification task). A series of consecutive works alleviated these limitations. 
For example, \citet{ziser-reichart-2017-neural} used a feed-forward neural network to learn the mapping, also exploiting the semantic similarity between pivots. Later \cite{ziser-reichart-2018-pivot, ziser-reichart-2019-task} proposed PBLM, a pivot-based language model, which also exploits the structure of the input text . Recently, \citet{ben-david-etal-2020-perl} integrated these ideas into the BERT architecture. They changed its Masked Language Modeling task so that pivots are more often masked than non-pivots, and the model should predict if a token is a pivot or not, and then identify the token only in the former case.


Despite the great success of this line of work on unsupervised DA for sentiment classification, the adaptation of the proposed ideas to sequence tagging tasks with cross-domain category shift is challenging. In this paper we solve this challenge and demonstrate that our solution yields state-of-the-art results on unsupervised DA for AE. We next survey existing work on this problem.

\subsection{Domain Adaptation for AE}
\label{DA_for_AE}

Like most recent DA works, the shared representation approach is prominent in DA for AE, where previous works align the different domains using syntactic patterns, reasoning that such patterns are robust across domains. For example, \citet{ding2017recurrent} learn a shared representation by training an RNN to predict rule-based syntactic patterns, populated by a pre-defined sentiment lexicon. While dependency relations-based rules are known to improve in-domain aspect extraction \cite{qiu-etal-2011-opinion,wang-etal-2016-recursive}, using hand-crafted patterns with a pre-defined sentiment lexicon heavily relies on prior knowledge and might not be robust when adapting to new, more challenging domains.
%
%
%
%
%

Similarly, inspired by the pivot-based modeling approach of \citet{blitzer2006domain}, \citet{wang-pan-2018-recursive} train a recursive recurrent network to predict source and target  dependency trees (obtained by an off-the-shelf parser \cite{klein-manning-2003-accurate}). Then, they jointly train the model to predict aspect and opinion words. Likewise, \citet{pereg2020syntactically} incorporate syntactic knowledge from an external parser \cite{dozat2017deep} into BERT via its self-attention mechanism. Relying on supervised parsers, these approaches naturally suffer from the degradation of such parsers when applied to resource-poor domains (e.g., user-generated content) or languages. Moreover, the work of \citet{wang-pan-2018-recursive} requires  additional human annotation for opinion word labels, which might not be available for new domains. 
%

\citet{li-etal-2019-transferable} avoids the need for external resources (except from an opinion lexicon), by applying a dual memory mechanism combined with a gradient reversal layer \cite{ganin2016domain}, in a model that jointly learns to predict aspect and opinion terms.  Recently, \citet{gong-etal-2020-unified} presented the Unified
Domain Adaptation (UDA) approach, the first to apply a pre-trained language encoder (BERT) to our task. Particularly, they apply self-supervised POS and dependency relation information as an auxiliary training task in order
to bias BERT towards
domain-invariant representations. Then, they apply instance re-weighting, and this way they perform DA at both the representation and the training instance level.

While syntactic pivot-based models contributed to unsupervised DA for AE, they do not harness the semantic properties of the involved domains. Moreover, some previous work rely on
external, resource-intensive syntactic models and on manual rules. Finally, the only previous work that applies a pre-trained language encoder (BERT) also focuses on syntax-driven adaptation. Our approach exploits the power of BERT for learning a cross-domain shared representation, but with semantically-driven, self-supervised pre-training tasks.

%% file: method.tex
\section{Domain Adaptation with DILBERT}

In this section, we introduce DILBERT, a new fine-tuning scheme for BERT \cite{devlin-etal-2019-bert}, customized for domain and category shift. Recall that BERT performs two pre-training tasks: (a) Masked Language Modeling (MLM), where some of the input tokens are randomly masked and the model should predict them based on their context; and (b) Next Sentence Prediction (NSP), where the model is provided with sentence pairs from its training data and it should predict whether one sentence is indeed followed by the other. 

DILBERT modifies the MLM task and presents a new pre-training task. Notice that DILBERT is applied to a BERT model that has already been trained on general text -- text that is not directly related to the source and target domains of interest -- with the standard MLM and NSP tasks. Hence, DILBERT can be seen as a fine-tuning step on the unlabeled data from the source and target domains. After DILBERT is applied, a task classifier is added on top of the resulting BERT model and this model is fine-tuned on the labeled source domain data to perform that aspect extraction task. This final model will eventually be applied to test data from the target domain. 
We next describe the two pre-training tasks of DILBERT.





\begin{figure*}[t!]
   \centering
   \includegraphics[width=6in]{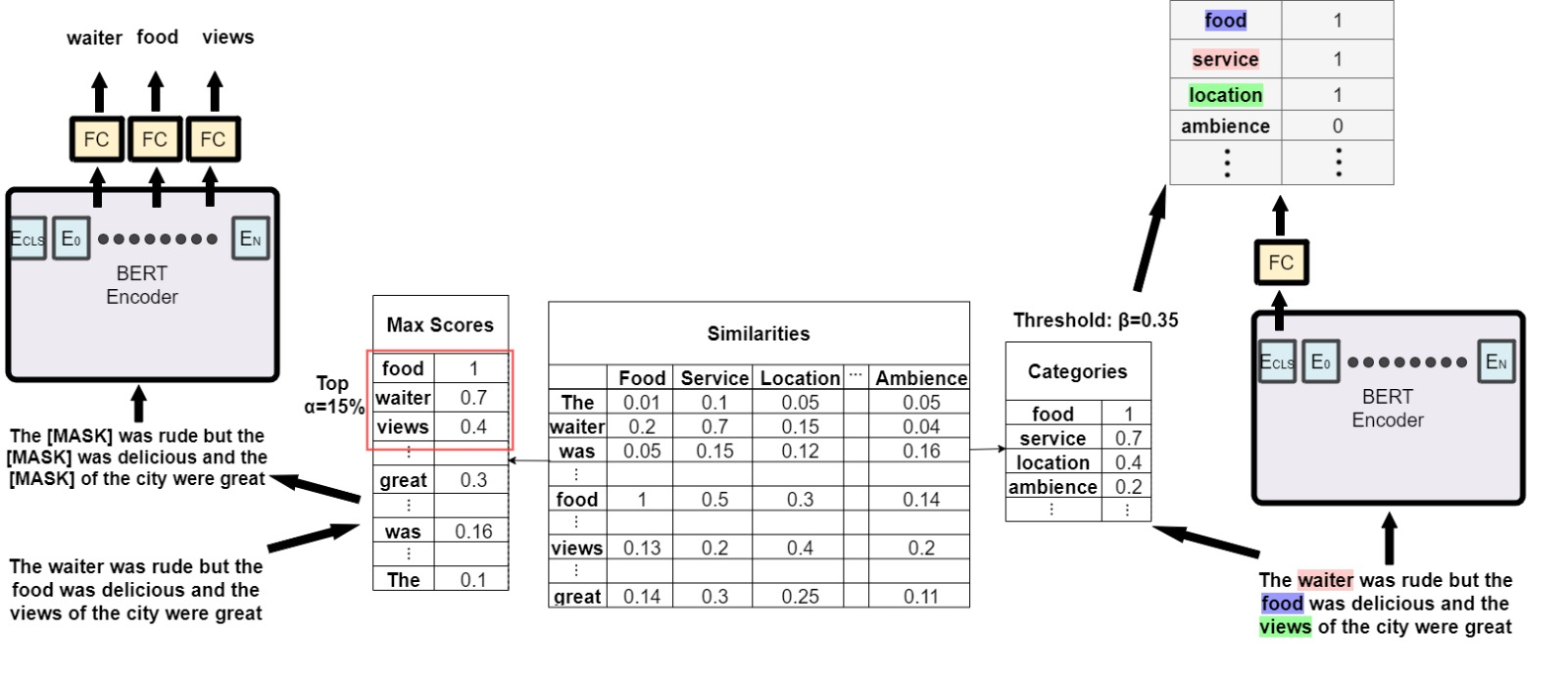}
   \caption{Illustrations of DILBERT fine-tuning tasks, CMLM and CPP, for an example input sentence.
First, we calculate a similarity score for each (word, category) pair using their pre-trained, static word embeddings. Then, for each word and for each category, we keep only the maximal score. For CMLM (left), we mask the top $\alpha$\% words, according to their score. For CPP (right), we construct a label vector in which the i\textsuperscript{th} coordinate is set to 1 if the i\textsuperscript{th} category score is greater than $\beta$, a threshold hyper-parameter, and to 0 otherwise. FC is a fully connected layer. The tasks are trained sequentially: First CMLM and then CPP.}
   \label{fig:DILBERT-overview}
\end{figure*}

\paragraph{Category-based Masked Language Modeling (CMLM)}

As noted above, \citet{ben-david-etal-2020-perl} integrated pivot-based training into the MLM pre-training task of BERT, in order to facilitate DA for sentiment classification. As noted in \S \ref{sec:intro}, it is challenging to define pivot features for the AE task, both because (a) it is a sequence-tagging, token-level task, and because (b) the aspect words and their categories tend to change across domains. Both these properties challenge the ``high MI with the task label" criterion in the pivot definition (criterion (b) of \S~\ref{sec:intro}), as this criterion is designed for (1) sentence (or larger text) labels where words can correlate with the task label (and hence property (a) above is challenging); and where (2) the correlation with the task label in the source domain is indicative of the correlation in the target domain (and hence property (b) is challenging).

To alleviate these limitations, we present the \textit{Category-based Masked Language Model (CMLM)} task: A variant of the BERT MLM pre-training task (left part of Figure \ref{fig:DILBERT-overview}). CMLM operates similarly to MLM, but with one difference: While in standard MLM the masked tokens are randomly chosen, CMLM harnesses information about the aspect categories in the source and target domains in order to mask words that are more likely to bridge the gap between the domains.\footnote{The aspect categories in the domains are part of the task definition, as aspect words or phrases must belong to one of these categories. For more details, see the SemEval annotation guide, section 2.1 in  \url{https://alt.qcri.org/semeval2016/task5/data/uploads/absa2016_annotationguidelines.pdf}. Interestingly, previous work did not use this information in their algorithmic solutions.}



We start by training static (non-contextualized) word embeddings on unlabeled data from both the source and the target domains. Then, we iterate over the input text and compute the cosine similarity between each input word and the word embedding of each of the aspect categories from both the source and the target.\footnote{This is the static word embedding of the category name.} For each word, we keep only the highest similarity score. Once we assigned a score to each input word, we mask the top $\alpha$\% of the input words (where $\alpha$ is a hyper-parameter).\footnote{We also considered schemes where a randomly selected sub-set of this list is masked, but observed no improvements.}

This masking mechanism ensures that the representation model focuses on words which can be viewed as proxy-pivots: Words that are likely to be aspect words in one of the domains. For example, given the input sentence \textit{The pasta was delicious, I really liked it.} from the Restaurants domain, CMLM would likely mask the word \textit{pasta}, which has a high similarity score with the \textit{Food} aspect category, where a vanilla MLM would randomly choose  a word to mask. 

\paragraph{Category Proxy Prediction (CPP)}

The CMLM task is about the prediction of masked words rather than their categories. While the masked words are strongly associated with categories, in some cases the coarser grained information that the category is represented in the text is most useful. To make our representation more category-informed, we add a second task: \textit{Category Proxy Prediction (CPP)} (right part of Figure \ref{fig:DILBERT-overview}). As implied by its name, this task is about predicting the aspect categories that are represented in the input text. However, in the representation learning phase of unsupervised DA, the unlabeled data from both domains does not have gold-standard labels of the aspect categories. We hence turn to proxy category labels instead.


More specifically, similarly to CMLM, we compute the cosine similarity between each input word and the (source and target) category names. This time, however, we keep for each category its highest score. Then we construct a binary vector, where each coordinate corresponds to one of the categories and its value is 1 if the score of that category is higher than $\beta$ and 0 otherwise. Here again $\beta$ is a hyper-parameter of the algorithm. Cross-entropy (which is also the loss function of CMLM) is a very natural loss for this task, as we are interested in a model that assigns high probabilities to aspect categories that are represented in the text and low probabilities to aspect categories that are not.


%% file: exp_setup.tex
\section{Experimental Setup}
\label{sec:expertiments}


\paragraph{Task and Domains}

We experiment with the task of cross-domain AE \cite{jakob2010extracting}. 
We consider data from the Amazon
laptop reviews (L) and the Yelp restaurant reviews (R) domains.  
The labeled data from the L domain is taken from the SemEval 2014 task on ABSA \cite{pontiki-etal-2014-semeval}. We follow \citet{gong-etal-2020-unified} and combine the SemEval 2014, 2015 and 2016 restaurant datasets \cite{pontiki-etal-2014-semeval, pontiki-etal-2015-semeval, pontiki-etal-2016-semeval} into a single  dataset, removing all duplicated examples. This results in 3045/800 and 3845/2158 train/validation examples for the L and R domains, respectively. At test time the target domain validation set serves as a test set. We obtain the list of category names for each domain from \citet{pontiki-etal-2016-semeval}.\footnote{The category list of each domain is provided in the appendix.} We used 863,000 laptop reviews from the Amazon reviews dataset of \citet{mcauley2015image}, and 570,000 reviews from the Yelp open dataset \footnote{https://www.yelp.com/dataset} as our unlabeled dataset for both domains, respectively.

To consider a more challenging setup, we experiment with the MAMS dataset \cite{jiang-etal-2019-challenge}, consisting of 5297 labeled reviews from the restaurant domain (M). We used 4297 reviews as a training set and 1000 as a validation set, and the same unlabeled data as for the R domain. Each review in the MAMS dataset has at least two aspects with different sentiment polarities, making it harder to adapt to, as the label distribution is different from that of the SemEval datasets. We consider adaptation from the L to the M domain and vice versa, adding two source-target pairs to our experiments (the M and R domains both address the same topic, and we consider them too similar for adaptation).  

\paragraph{Experimental Protocol}

Focusing on unsupervised DA, we have access to unlabeled data from the source and target domains and labeled data from the source domain only. Following the protocol of representation learning for DA (\S~\ref{sec:previous}), we learn a domain-invariant representation model using the unlabeled data from both domains. Then, we use the source domain labeled data to fine-tune the representation model on the downstream task. This model is eventually applied to the target domain test set data.  

For DILBERT, our representation learning phase consists of two fine-tuning tasks. First, we use the large unlabeled reviews data from both domains to further train a BERT model on the CMLM task. In all our experiments (for DILBERT and the baselines) we employ the BERT-Base-Uncased model of the Hugging-Face \cite{Wolf2019HuggingFacesTS} transformers package.\footnote{https://github.com/huggingface/transformers} Following \citet{devlin-etal-2019-bert}, we fine-tune all of BERT model layers. For the CMLM task, we mask full words instead of the default sub-word masking, i.e., if we choose to mask a word, all its corresponding sub-words are masked. 

Then, we use target unlabeled data to further fine-tune on the CPP task. We add a linear layer on top of the [CLS] token, and then  feed forward the outputs through a sigmoid layer, using the sum of per-category binary cross-entropy losses. Once the representation learning phase is completed, we remove the CPP additional layer and the CMLM sequence classification heads. Then, we add a logistic regression head on top of all the word-level outputs and fine-tune the model on the source labeled data for the aspect. Finally, we use the fine-tuned model for predictions on the target domain test set.

To compute the similarity between words and aspect categories, we experiment with two word embedding sets: (a) The fastText  pre-trained word embeddings (we refer to the DILBERT model that uses these embeddings as DILBERT-PT-WE);\footnote{https://github.com/facebookresearch/fastText/} and (b) FastText word embeddings trained on the unlabeled data from both domains (DILBERT-CT-WE; with a learning rate of 5e-5, a batch size of 4, and one training epoch.). We report results with DILBERT-CT-WE as they are better (see \S~\ref{sec:Scare_Unlabeled}).

For all the baselines, we keep the same protocol and design choices, unless otherwise stated. 

\paragraph{Baselines}

The first baseline is UDA \cite{gong-etal-2020-unified},\footnote{We use the code from the author's GitHub: https://github.com/NUSTM/BERT-UDA.} a pre-trained transformer-based neural network that performs pre-training using syntactic-driven auxiliary tasks, combined with an adversarial component (\S~\ref{sec:previous}). There are two variants of UDA, differing with respect to their initialization. UDA-BASE (UDA-B) is initialized to the BERT-BASE model, while UDA-EXTENDED (UDA-E) is initialized with a BERT model which was further fine-tuned with over 20GB from the Yelp Challenge dataset and the Amazon Electronics dataset \cite{xu2019bert}. As shown in \citet{gong-etal-2020-unified}, UDA-E outperforms all previous cross-domain AE work by a large margin (\S~\ref{DA_for_AE}), and hence UDA-E and UDA-B are the previous works we compare to.
 
 To better understand the effect of our customized pre-training method, we also compare our model to a variant where everything is kept fixed except that the fine-tuning stage on the source and  target unlabeled data is performed with the standard BERT model rather with DILBERT (BERT-S\&T). We further compare to two similar variants that reflect a condition where we do not have access to target domain data (no domain adaptation (No-DA) setups): BERT-S, where the fine-tuning stage is performed with the standard BERT model and on unlabeled source domain data only, and Vanilla, where no unlabeled data fine-tuning is performed.
  
 %
 %
 
Finally, in order to understand the relative importance of the two DILBERT pre-training tasks, we experiment with the D-CMLM model (where unlabeled data fine-tuning is done with DILBERT, but only with the CMLM task), and with the D-CPP model  (where unlabeled data fine-tuning is done with DILBERT, but only with the CPP task). We also evaluate the performance of an in-domain classifier (BERT-ID) -- i.e., our classifier when trained on the target domain training set and evaluated on the target domain test set. This model can be seen as an upper bound on the performance we can realistically hope a DA model to achieve.

\paragraph{Hyper-parameter Tuning}
\label{sec:hyper}

All experiments are repeated five times using different random seeds and the average results  are reported.  
As stated, all models are based on the HuggingFace BERT-base Uncased pre-trained model. 
All models were tuned on the same set of hyper-parameters and the same data splits.  The validation examples, used for hyper-parameter tuning, are from the source domain.\footnote{Full details about the hyper-parameter tuning, best hyper-parameter configurations, as well as computing infrastructure and run-time details are in Appendix \ref{app:computing} and \ref{app:hyper}.}

%

%% file: results.tex
\section{Results and Analysis}


\begin{table}[t!]
\small 
	\centering
	\begin{tabular}{|l|l|l|l|l|l|}
\hline
	& \textbf{M-L}  & \textbf{L-M}   & \textbf{R-L}   & \textbf{L-R}    & \textbf{Average} \\ \hline
\multicolumn{6}{ |c| }{DILBERT Methods} \\
\hline
		\textbf{DILBERT}   & \textbf{43.72} & \textbf{58.96}    & \textbf{56.07}    & \textbf{61.04}   & \textbf{54.95}   \\ \hline
        \textbf{D-CMLM}    & 41.97 & 52.74    & 55.48    &   59.36 & 52.39      \\ \hline
             
\hline   

        \textbf{D-CPP}    & 31.08 & 29.06    & 36.16    &  40.55  &  34.21     \\ \hline

        		\multicolumn{6}{ |c| }{Previous Baselines} \\
\hline
		\textbf{UDA-E}    & 41.29 & 45.62    & 53.51    &   56.12 & 49.14     \\ \hline
				\textbf{UDA-B}    & 36.52 & 39.59    & 48.35    &   49.52 & 43.50     \\ \hline
					\textbf{BERT-S\&T}    & 34.30 & 29.82    & 45.89    &   39.92 & 37.48     \\ \hline

			\multicolumn{6}{ |c| }{No-DA} \\
\hline
\textbf{BERT-S}    & 28.9 & 27.53    & 36.35    &   44.12 & 34.22 \\
\hline
\textbf{Vanilla}    & 32.21 & 22.7    & 32.12   &   34.25 & 30.32 \\
\hline
	\end{tabular}

	\begin{tabular}{|l|l|l|l|l|}
		\hline
			\multicolumn{5}{ |c| }{In Domain} \\
			\hline
	& \textbf{M}  & \textbf{L}   & \textbf{R}     & \textbf{Average} \\ \hline

\textbf{BERT-ID}    & 77.87 & 83.8    & 78.82 &    80.16 \\
\hline
	\end{tabular}
	\centering

	\caption{F1 results for both DA (top label) and in-domain (bottom table, provided for reference) setups. In the top table, column titles correspond to source-target domain pairs.} 
	\label{tab:main}
\end{table}

\begin{table*} [h!]
\centering
\scriptsize
\subfloat[Subtable 1 list of tables text][\label{cmlm-a}\textit{There was only one \textcolor{blue}{MASK} for the whole restaurant upstairs}. The masked word is \textcolor{blue}{\textit{waiter}}.]
{
\begin{tabular}{|l|l|l| ccc }

\hline
	 \textbf{BERT-S\&T}  & \textbf{D-CMLM}  \\ \hline
room    & server   \\
\hline
window    & waiter   \\
\hline
table    & bartender   \\
\hline
seat    & waitress   \\
\hline
person    & table   \\
\hline
\end{tabular}}
\qquad
\subfloat[Subtable 2 list of tables text][\label{cmlm-b}\textit{\textcolor{blue}{MASK} was just enough for me , but may not be for a big eater}. The masked word is \textcolor{blue}{\textit{portions}}.]{
\begin{tabular}{|l|l|l| ccc}

\hline
	 \textbf{BERT-S\&T}  & \textbf{D-CMLM}  \\ \hline
that    & portion   \\
\hline
size    & salad   \\
\hline
it    & bowl   \\
\hline
this    & rice   \\
\hline
which    & sandwich   \\
\hline
\end{tabular}}
\qquad
\subfloat[Subtable 2 list of tables text][\label{cmlm-c}\textit{\textcolor{blue}{MASK1 MASK2} would not fix the problem unless i bought your plan for \$ 150 plus}. The masked words are \textcolor{blue}{\textit{tech support}}.]{
\begin{tabular}{|l|l|l|l|l| ccc}
\hline
\multicolumn{2}{|c|}{MASK1} & \multicolumn{2}{|c|}{MASK2} \\
\hline
	 \textbf{BERT-S\&T}  & \textbf{D-CMLM} &  \textbf{BERT-S\&T} & \textbf{D-CMLM} \\ \hline
my    & costumer & they & support   \\
\hline
you   & dell & i& r   \\
\hline
this    & tech &you & y   \\
\hline
i    & ace & it & ware   \\
\hline
you    & warrant &plan &us   \\
\hline
\end{tabular}}
\caption{Top 5 predictions for a given masked word according to BERT-S\&T
and D-CMLM, when trained on unlabeled data from the R and L domains. Examples (a)
and (b) are taken from the R domain test set, and (c) from the L domain test set. For this analysis, BERT-S\&T
and D-CMLM were not fine-tuned for the AE task.}
\label{tab:qual}
\end{table*}

\paragraph{Main Results} 

Table~\ref{tab:main} presents our results. As in previous work, we report the exact-match F1 score over aspect words and phrases. It shows that DILBERT, our customized pre-training procedure, outperforms all other alternatives across all setups. DILBERT reduces the error of the best non-DILBERT baseline, UDA-E, by over 5\% on average while using a fraction of the unlabeled data (recall that UDA-E employs a BERT model that is heavily pre-trained with data from the Restaurants and Electronic domains). The performance gap between DILBERT and UDA-E is even larger when considering the most challenging L-M setup, with over 13\% improvement. The performance gaps from UDA-B, which like DILBERT is initialized with a BERT-base model, are much larger (11.45\% on average across settings, 19.37\% for L-M).

The comparison to the DILBERT variants that perform only one of its pre-training tasks provides a clear picture of their relative importance. Clearly, D-CMLM, the DILBERT variant which performs only the CMLM task, is an effective model, and is outperformed only by the full DILBERT. Note, however, that while the average performance gap between these models is 2.56\%, in the L-M setup it is as high as 6.22\%. While D-CPP is not competitive as a standalone model, its combination with D-CMLM (to form the full DILBERT model) consistently improves D-CMLM, in all DA settings. The comparison to BERT-S\&T (where the representation learning stage is performed with the standard BERT rather than with DILBERT), indicates the great overall impact of DILBERT with both its tasks, as the average improvement of DILBERT is as high as 17.47\% on average and 29.14\% on L-M.

The performance of the No-DA baselines confirms our intuition that the MAMS restaurant domain is more challenging from a DA perspective compared to the SemEval restaurant domain (at least when adapting form/to the SemEval L domain). Additionally, and not surprisingly, BERT-S\&T, which is trained on unlabeled data from both domains, outperforms BERT-S which is exposed only to source domain data. 

Finally, a comparison to BERT-ID, the in-domain model which is trained and tested in the target domain, provides an indication about the performance gap that is yet to be closed. It also provides an indication of the error reduction (ER) that has already been achieved. Comparing the average performance of the Vanilla No-DA classifier to that of BERT-ID (the cross-domain error) and to that of DILBERT reveals that DILBERT cuts 24.63\% from  an error of 49.84\% -- an ER of 49.41\% (for comparison, the ER of UDA-E is 37.76\%).

\paragraph{The Limited Unlabeled Data Scenario}
\label{sec:Scare_Unlabeled}

\begin{figure}[h!]
\centering  
\begin{subfigure}[b]{\columnwidth} 
\includegraphics[width=\columnwidth,height=0.16\textheight]{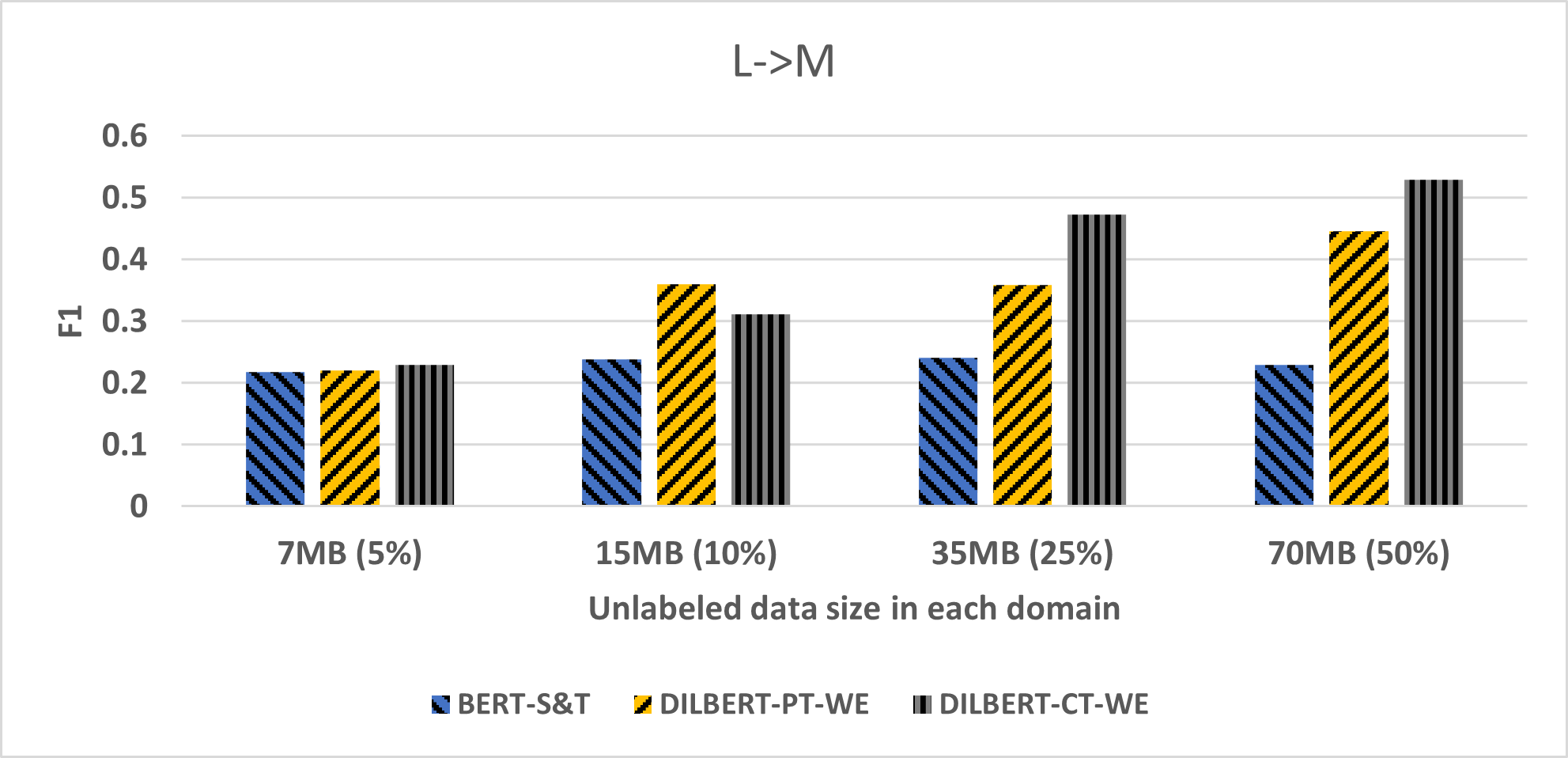}
\end{subfigure}
\vspace{0.2cm}
\begin{subfigure}[b]{\columnwidth} 
\includegraphics[width=\columnwidth,height=0.16\textheight]{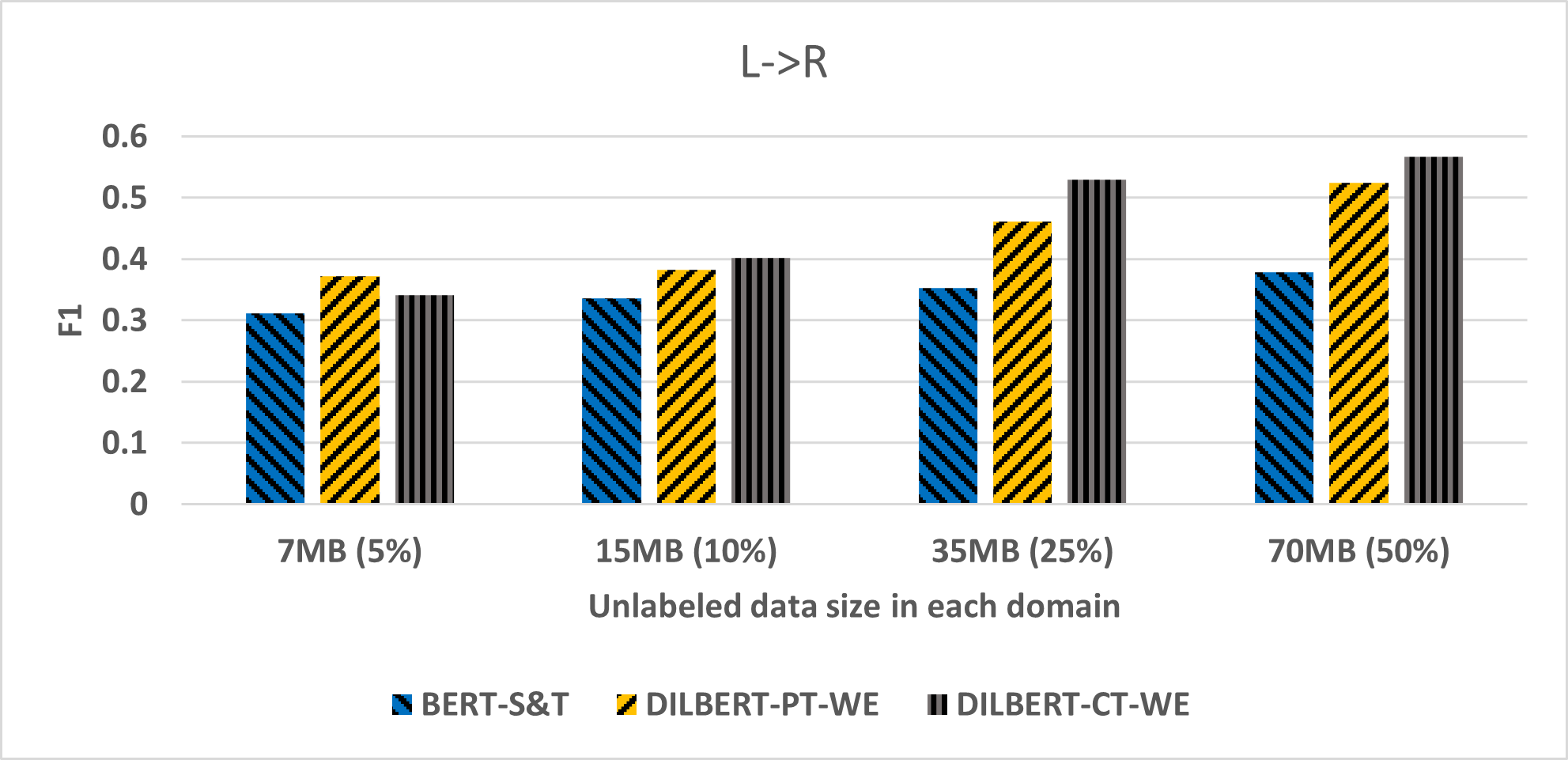}
\end{subfigure}
\caption{\normalsize AE performance as a function of the unlabeled dataset size used for representation learning.}
\label{fig:low-resource}
\end{figure}

While the main bottleneck of adapting to a new domain is the lack of labeled data, obtaining large amounts of unlabeled data is challenging for truly resource-poor domains or languages, and may not be sufficient to learn domain-invariant representations \cite{ziser-reichart-2018-deep}. To simulate such a scenario, we fed the DILBERT-CT-WE and DILBERT-PT-WE models (see \S~\ref{sec:expertiments}) as well as the BERT-S\&T baseline with randomly sampled  unlabeled data subsets from the source and the target.
%
%
Figure ~\ref{fig:low-resource}  demonstrates that the two DILBERT variants outperform BERT-S\&T when unlabeled data is scarce, and in some cases, DILBERT outperforms BERT-S\&T even when using less unlabeled data (e.g., DILBERT with 35MB of unlabeled data from each domain, compared to BERT-S\&T with 70MB). 


\paragraph{CMLM Probing}

We would next like to shed more light on the quality of the CMLM task, which has the strongest impact on DILBERT's results. For this aim, we fine-tune the D-CMLM and BERT-S\&T models on the unlabeled data of the R and L domains, and apply them to test sentences from these domains, without task-related (AE) fine-tuning with label data. Table~\ref{tab:qual} presents the words predicted by these models for three representative masking tasks. Obviously, the predictions of D-CMLM are much more semantically related to the masked tokens. While this is a qualitative analysis, limited in nature to a small number of examples, our manual inspection of the results suggests that this pattern is the rule rather than the exception.

\paragraph{Generalizing Beyond Category Names}

We would further like to verify that by considering the information encoded in the category names DILBERT can better identify aspect words and phrases that are not identical to one of these category names. We hence re-evaluate all models such that aspect words that are also category names (e.g., in the restaurant review sentence: \textit{The food was great} where \textit{food} is both an aspect word and the name of its category) are not considered in the evaluation.\footnote{For more details about this scenario -- the category names and their prevalence among the aspect terms in the texts, see appendix~\ref{aspect-cat}.} 

Table \ref{tab:no-categories} reports the results of this evaluation. The observed patterns are similar to the main evaluation results: DILBERT is still the best DA model by a large margin. Not surprisingly, the absolute results of all models are lower than in the main evaluation as the excluded aspect words are more typical and hence easier for the models to identify.
\begin{table}
\small 
	\centering
	\begin{tabular}{|l|l|l|l|l|l|}
\hline
	& \textbf{M-L}  & \textbf{L-M}   & \textbf{R-L}   & \textbf{L-R}    & \textbf{Average} \\ \hline
\multicolumn{6}{ |c| }{DILBERT Methods} \\
\hline
		\textbf{DILBERT}   & \textbf{36.18} & \textbf{48.35} & \textbf{48.69}    & \textbf{43.81}   & \textbf{44.26}   \\ \hline
        \textbf{D-CMLM}    & 35.34 & 46.21    & 48.18    &   43.32 & 43.26      \\ \hline
             
\hline   

        \textbf{D-CPP}    & 23.06 & 16.94    & 31.22    &  27.72  &  24.74     \\ \hline

        		\multicolumn{6}{ |c| }{Previous Baselines} \\
\hline
		\textbf{UDA-E}    & 32.74 & 37.37    & 46.13    &   39.54 & 38.95     \\ \hline
				\textbf{UDA-B}    & 30.73 & 29.71    & 38.06    &   32.38 & 32.72     \\ \hline
					\textbf{BERT-S\&T}    & 25.87 & 22.78    & 35.77    &   31.35 & 28.94     \\ \hline

			\multicolumn{6}{ |c| }{No-DA} \\
\hline
\textbf{BERT-S}    & 21.43 & 18.14    & 28.85    &   29.03 & 24.36 \\
\hline
\textbf{Vanilla}    & 25.45 & 18.76    & 29.14   &   24.79 & 24.53 \\
\hline
	\end{tabular}

	\begin{tabular}{|l|l|l|l|l|}
		\hline
			\multicolumn{5}{ |c| }{In Domain} \\
			\hline
	& \textbf{M}  & \textbf{L}   & \textbf{R}     & \textbf{Average} \\ \hline

\textbf{BERT-ID}    & 69.7 & 75.19    & 67.48 &   70.79 \\
\hline
	\end{tabular}
	\centering

	\caption{The results of a re-evaluation of all the models where aspect words that are identical to category names are excluded from the evaluation.} 
	\label{tab:no-categories}
\end{table}

\paragraph{Standard Deviations}

Finally, Table \ref{tab:std} reports the standard deviations for all models across the five folds of the cross-validation protocol. DILBERT has the lowest averaged standard-deviation among all DA models and No-DA baselines. This is obviously another advantage of our proposed model, particularly that in unsupervised domain adaptation there is no labeled target domain data available for model selection, and hence stability to random seeds is crucial \cite{ziser-reichart-2019-task}.

\begin{table}
\small 
	\centering
	\begin{tabular}{|l|l|l|l|l|l|}
\hline
	& \textbf{M-L}  & \textbf{L-M}   & \textbf{R-L}   & \textbf{L-R}    & \textbf{Average} \\ \hline
\multicolumn{6}{ |c| }{DILBERT Methods} \\
\hline
		\textbf{DILBERT}   & 1 & \textbf{0.4}    & 1.3    & 1.8   & \textbf{1.1}   \\ \hline
        \textbf{D-CMLM}    & 2 & 0.95    & \textbf{1.1}    &   \textbf{1} & 1.26      \\ \hline
             
\hline   

        \textbf{D-CPP}    & \textbf{0.62} & 5.03    & 2.84    &  3.95  &  3.11     \\ \hline

        		\multicolumn{6}{ |c| }{Previous Baselines} \\
\hline
		\textbf{UDA-E}    & 1.03 & 1.85    & 0.6    &   2.17 & 1.41     \\ \hline
				\textbf{UDA-B}    & 1.51 & 1.68    & 1.54    &   2.29 & 1.76     \\ \hline
					\textbf{BERT-S\&T}    & 2.08 & 4.3    & 2.17    &   1.39 & 1.85     \\ \hline

			\multicolumn{6}{ |c| }{No-DA} \\
\hline
\textbf{BERT-S}    & 1.45 & 2.39    & 2.17    &   1.39 & 1.85 \\
\hline
\textbf{Vanilla}    & 1.23 & 2.63    & 2.2   &   1.47 & 1.88 \\
\hline
	\end{tabular}

	\begin{tabular}{|l|l|l|l|l|}
		\hline
			\multicolumn{5}{ |c| }{In Domain} \\
			\hline
	& \textbf{M}  & \textbf{L}   & \textbf{R}     & \textbf{Average} \\ \hline

\textbf{BERT-ID}    & 0.33 & 0.77    & 0.41 &    0.51 \\
\hline
	\end{tabular}
	\centering

	\caption{Standard deviations for all  models across the five folds of the cross-validation protocol.} 
	\label{tab:std}
\end{table}



%% file: conclusions.tex
\section{Conclusions and Future Work}

We have presented DILBERT, a customized pre-training approach for unsupervised DA with category shift, and apply it to the task of aspect extraction. We demonstrate that by fine-tuning with a modified version of the BERT  pre-training tasks, we can better adapt to new domains and aspect categories, even in resource-poor scenarios where unlabeled data is limited. To make our experimentation more challenging, we presented a new AE domain (M), which is substantially different from previously presented ones.  

In future work, we would like to extend our approach so that it can jointly solve the  aspect extraction and sentiment analysis tasks (the ABSA task). Moreover, we would like to verify the quality of our approach in additional domains, tasks (i.e., going beyond AE to other tasks that present category shifts when domains change) and eventually even languages. 

%% file: supp-materail.tex
\appendix
\section{Appendix}
\subsection{Aspect Categories}
\label{aspect-cat}

Table \ref{tab:categories} provides the category names of our three domains, which were available to our model. There are 28 categories for the laptops (L) domain and 9 categories for the restaurants domains (R and M). In the R test set there are 876 unique aspect terms (i.e., unique words or phrases that are annotated as aspects), and 21.1\% of these are words that are identical to one of the category names.
In the M test set the corresponding numbers are 835 unique aspect terms, and 12.4\% of these are words that are identical to one of the category names. In the laptops (L) test set the corresponding numbers are 387 unique aspect terms, and 12.7\% of these are words that are identical to one of the category names.

\begin{table}
\small 
	\centering
	\begin{tabular}{|l|l|}
\hline
	 \textbf{Laptops}  & \textbf{Restaurants}     \\  
\hline
		   cooling, multimedia,warranty, & drinks, food, service, \\OS, portability, motherboard, & prices, style, location, \\support, keyboard, quality, & ambience, quality, \\powersupply, ports, fan,& restaurant \\software, memory, battery,& 
		   \\graphics, display, usability,& \\price, design,  connectivity,& \\performance, shipping, laptop,& \\disk, hdd, cpu, mouse&  \\  
\hline
	\end{tabular}
	\centering
	\caption{The list of aspect categories of each of the participating domains. \textit{Restaurants} refer both to the R and the M domains.} 
	\label{tab:categories}
\end{table}

\subsection{Computing Infrastructure, Run-time and Number of Parameters}
\label{app:computing}

All models were trained on a Quadro RTX 6000 machine, using 1 GPU card.
The average run-time for the CMLM training was about 2 hours.
The average run-time for the CPP training was about 6 minutes.
The average run-time for the AE training was about 6 minutes.
The inference time on the test set for the AE task was about 1 minute.
The BERT-base model has 109.5M parameters.
The CMLM architecture has additional 620K parameters.
The CPP architecture has additional 21.5K parameters.
The AE task classifier has additional 2307 parameters.

\subsection {Hyper-parameter Tuning}
\label{app:hyper}

All  models are based on the HuggingFace BERT-base Uncased pre-trained model.  We use their default word-piece vocabulary, and the AdamW optimizer \cite{loshchilov2019decoupled}  with an $\epsilon = 1e-8$ and a linearly decreasing learning rate schedule. 

The number of words to mask ($\alpha$) in the CMLM task was chosen among 5\%, 10\%, and 15\% of the review length.
For the CPP task, the number of epochs was chosen among  \{1, 2, 3\}, and the $\beta$ threshold among $[0.25 ,0.251, \ldots ,0.45]$. The learning rate was 5e-5 and the batch size was 8. 

In the AE classification stage, the number of epochs was chosen among \{2,3,4\}, the batch size among \{16, 32, 64\} and the learning rate among \{2e-5, 3e-5, 5e-5\}.

For both UDA-B and UDA-E, the hyper-parameter search of \citet{gong-etal-2020-unified} is performed over a sub-set of the set we consider here. We hence re-run all of their experiments with our grid search, which led to better performance than reported in their work.

\paragraph{Best Hyper-parameters}

The hyper-parameters of the best DILBERT configuration across cross-validation folds and DA setups were as follows:

\begin {itemize}

\item CMLM, $\alpha$: 10\%.
\item CPP, number of epochs : 1.
\item CPP, $\beta$:\\ R-L: 0.353, L-R, L-M: 0.406, M-L: 0.362.
\item AE (task classifier), number of epochs: 3
\item AE (task classifier), batch size: 32.
\item AE (task classifier), learning rate: 5e-5.

\end {itemize}
